\documentclass{IEEEcsmag}

\usepackage{cite}
\usepackage{booktabs}
\usepackage{makecell}
\usepackage{multirow}

\usepackage[colorlinks,urlcolor=blue,linkcolor=blue,citecolor=blue]{hyperref}
\expandafter\def\expandafter\UrlBreaks\expandafter{\UrlBreaks\do\/\do\*\do\-\do\~\do\'\do\"\do\-}
\usepackage{upmath,color}

\jvol{XX}
\jnum{XX}
\paper{8}
\jmonth{Month}
\jname{Publication Name}
\jtitle{Publication Title}
\pubyear{2026}

\graphicspath{{fig/}}

\begin{document}

\title{Visualization of Machine Learning Models through Their Spatial and Temporal Listeners}

\author{Siyu Wu}
\affil{School of Computer Science \& Engineering, Beihang University, Beijing, 100191, China}

\author{{Lei} Shi*}
\affil{School of Computer Science \& Engineering, Beihang University, Beijing, 100191, China}

\author{{Lei} Xia}
\affil{School of Computer Science \& Engineering, Beihang University, Beijing, 100191, China}

\author{Cenyang Wu}
\affil{Institute of Medical Technology, Peking University Health Science Center and National Institute of Health Data Science, Peking University, Beijing, 100191, China}

\author{Zipeng Liu}
\affil{School of Software, Beihang University, Beijing, 100191, China}

\author{Yingchaojie Feng*}
\affil{National University of Singapore, Singapore}

\author{{Liang} Zhou*}
\affil{Institute of Medical Technology, Peking University Health Science Center and National Institute of Health Data Science, Peking University, Beijing, 100191, China}

\author{Wei Chen}
\affil{State Key Laboratory of CAD\&CG, Zhejiang University, Hangzhou, 310058, China}

\begin{abstract}
  \looseness-1
  Model visualization (ModelVis) has emerged as a major research direction, yet existing taxonomies are largely organized by data or tasks, making it difficult to treat models as first-class analysis objects. We present a model-centric two-stage framework that employs abstract listeners to capture spatial and temporal model behaviors, and then connects the translated model behavior data to the classical InfoVis pipeline. To apply the framework at scale, we build a retrieval-augmented human--large language model (LLM) extraction workflow and curate a corpus of 128 VIS/VAST ModelVis papers with 331 coded figures. Our analysis shows a dominant result-centric priority on visualizing model outcomes, quantitative/nominal data type, statistical charts, and performance evaluation. Citation-weighted trends further indicate that less frequent model-mechanism-oriented studies have disproportionately high impact while are less investigated recently.
  Overall, the framework is a general approach for comparing existing ModelVis systems and guiding possible future designs.
\end{abstract}

\maketitle

\begin{figure*}
  \centering
  \includegraphics[width=0.93\linewidth]{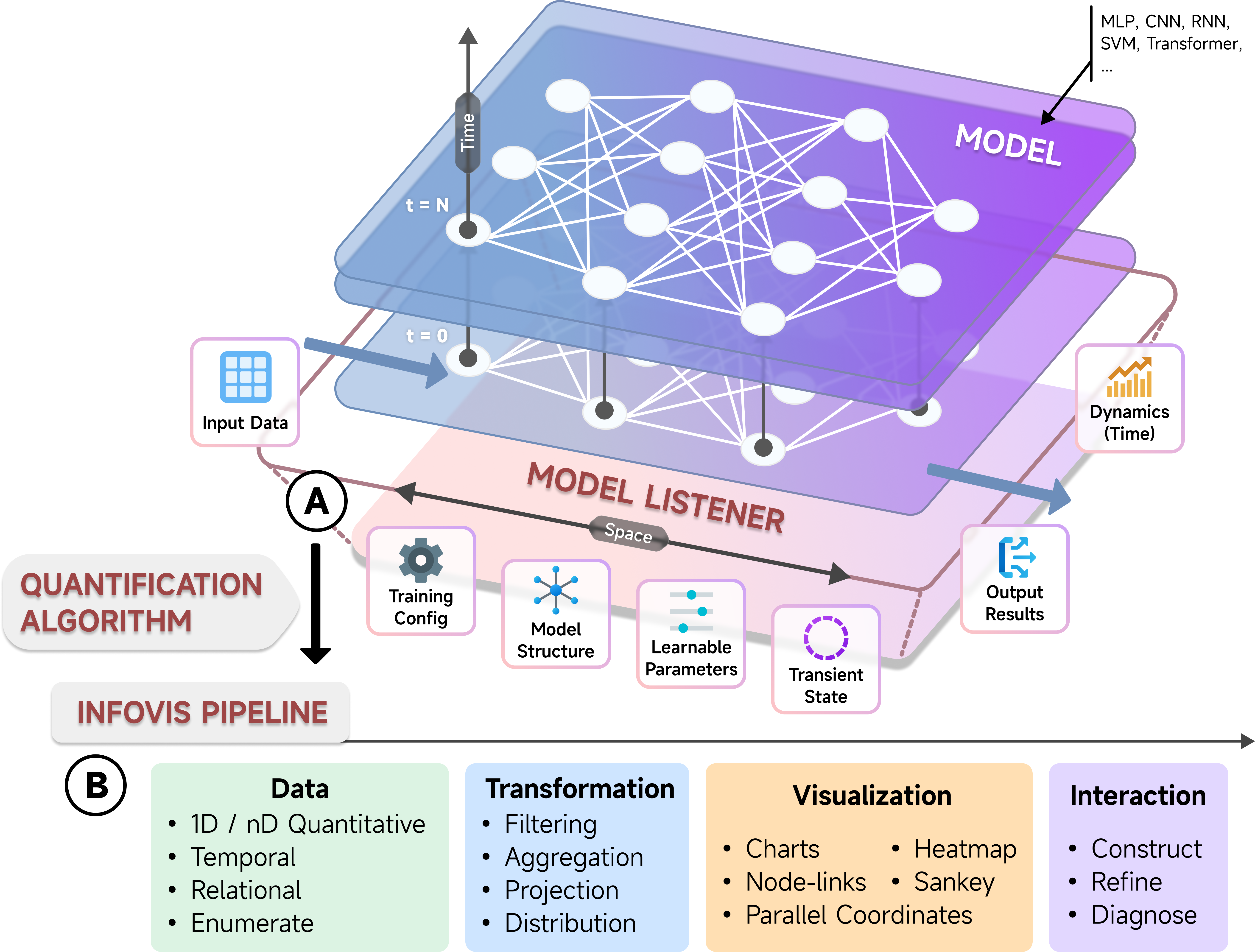}
  \caption{
    Overview of our model-centric ModelVis framework.
    (A) A model listener layer captures model evidence from \textit{input data}, \textit{training configuration}, \textit{model structure}, \textit{learnable parameters}, \textit{transient state}, \textit{dynamics (time)}, and \textit{output results}, and quantifies these observations into analysis-ready data.
    (B) The quantified outputs are then mapped to the classical InfoVis pipeline, including data organization, transformation, visualization, and interaction.
  }
  \label{fig:framework}
\end{figure*}

\chapteri{M}odel visualization (ModelVis)---the use of computer-generated, visual representation of mathematical models (instead of data) to gain insights---has emerged as a new area in visualization, data science, and artificial intelligence (AI).
Without the loss of generality, we refer to models as both predictive and descriptive ones, although the focus of the community is on the former, which is also known as machine learning models.
Numerous aliases exist for ModelVis and are used interchangeably: interpretable machine learning, explainability of machine learning models, XAI, open the black-box of models, etc.
The topic becomes important as models become black-boxes after the introduction of complex model-building algorithms such as deep learning.
Humans can hardly recognize the internal mechanism of models even though they have designed the training algorithm.
As more data is applied to train the model, the predictive power is increasingly being shaped by the underlying data, but not the human designer of the algorithm.
On the other hand, modern machine learning models, for example, LLMs, are huge in volume and evolve quickly, overwhelming any human being without appropriate visual guidance.

Visualization has been successfully applied to understand machine learning models.
Zeiler and Fergus leverage visualization and an interpretation method, namely, deconvolution, to win the ImageNet Challenge in 2013~\cite{zeiler2014visualizing}, right after the deep learning breakthrough by AlexNet in 2012. The TensorFlow Playground by Google~\cite{TensorFlowPlayground} introduces an interactive educational tool for learning the internal mechanism of basic multi-layer neural networks.
Users without prior knowledge of machine learning or even computer science can understand and diagnose deep learning models via the interactive visualization platform.
Eventually, as hundreds or more ModelVis techniques have been proposed, the community calls for ModelVis frameworks and taxonomies that can help to comprehend, compare, and apply existing methods according to their relevant usage scenarios and problem settings.

Taxonomies of ModelVis are available in surveys.
However, most existing taxonomies use visualization-oriented perspectives, e.g., by visualization goals
~\cite{Seifert2017-SOS-12, Liu2017-SOS-8, ChooLiu2018-SOS-2, Yu2018-SOS-14, Zhang2018-SOS-15}, information visualization (InfoVis) and visual analytics (VA) pipelines
~\cite{Lu2017-SOS-9, Lu2017-SOS-10, Wang2016-SOS-13, Zhang2018-SOS-15, Sacha2016-SOS-17},
visualization data and users \cite{Wang2024-SA2020-1, Dudley2018-SOS-3, Yu2018-SOS-14}.
Model-centric taxonomies are rare, although model is the primary subject in ModelVis, as opposed to data/visualization subject in InfoVis/VA. Therefore, our main motivation in this paper is to investigate the possibility of developing a model-centric taxonomy/framework that can encapsulate existing and future ModelVis techniques. Furthermore, we ask what are the similarities and dissimilarities of such a ModelVis framework compared to classical InfoVis/VA pipelines.

To this end, this work is initiated based on two observations.
First, machine learning models are complex systems composed of multiple components, e.g., input data (independent variables), model structure, learning algorithm, and output data (dependent variables). Meanwhile, InfoVis techniques are studied and designed according to the type of data to consume, for example, text, tables, or networks.
This introduces a mismatch where the key lies in that---models are typically not singular-typed data.
Second, to overcome this limitation, most existing ModelVis techniques translate machine learning models into separate types of data for visualization, though this could be unintentional, e.g., extracting the model structure as networks, computing the performance of models as 1D data, mapping and correlating input and output data into multi-dimensional data.
According to these observations, the main idea of this work is to introduce an abstract model listening layer (Figure~\ref{fig:framework}.A), which captures different aspects of a machine learning model for possible visualization.
Model quantification algorithms are deployed after the listening layer to further translate model behavior information into consumable data objects.
Subsequently, as the output data objects become the relayed new input, the classical InfoVis pipeline can be reused to serve the follow-up visualization and interaction tasks (Figure~\ref{fig:framework}.B).

With this new ModelVis framework, we present a literature analysis method and examine a large corpus of VA papers that could be related to the ModelVis topic.
The method features a three-stage workflow leveraging the synergy of humans and LLMs to extract relevant ModelVis dimensions in the new framework from these papers.

Post-hoc analysis on the LLM extraction result reveals the distributions and trends of ModelVis techniques and validates the feasibility of our new framework.

The main contributions of this paper are as follows.
\begin{itemize}
  \item A new model-centric taxonomy that categorizes ModelVis techniques primarily by their listening mechanism on model behavior, from both spatial and temporal dimensions. The framework extends the classical InfoVis pipeline by abstracting the essential model-data translation mechanism.
  \item An LLM-based literature analysis method that maps state-of-the-art ModelVis techniques onto the new framework without the need of full, costly human annotations, while still preserving high accuracy. On a core ModelVis literature collection, the mapping result validates the effectiveness of the proposed model-centric framework.
  \item The post-hoc analysis on framework/taxonomy mapping elicits several interesting results while suggesting potential limitations on the current ModelVis research. For example, the importance of multi-dimensional data visualization and commodity statistical charts in ModelVis, and the recent decline of investigation into deeper mechanism (e.g., model structure, transient states) and the temporal dynamics of machine learning models.
\end{itemize}

\section{RELATED WORK}
\label{sec:related-works}

We summarize related approaches on the taxonomy of model visualization.

According to the survey of surveys on ModelVis published in 2020~\cite{chatzimparmpas2020survey}, 18 noteworthy surveys or taxonomy-discussing papers on interpreting machine learning models are available.
We manually supplement with five more appearing after 2020.
Out of these 23 surveys, 16 explicitly propose at least one technical taxonomy of model visualization works.

The first group of surveys adopts the most popular categorization that classifies techniques according to their high-level visualization goals and tasks
~\cite{Seifert2017-SOS-12, Liu2017-SOS-8, ChooLiu2018-SOS-2, Yu2018-SOS-14, Zhang2018-SOS-15}.
A typical example is by Choo and Liu~\cite{ChooLiu2018-SOS-2}, which explicitly discusses three key goals of visual analytics for explainable deep learning: understanding, debugging/diagnosis, and refinement/steering. This categorization is followed in a large portion of related work on model visualization goals~\cite{Liu2017-SOS-8, Zhang2018-SOS-15}.

Another collection of surveys reports model visualization literature by their stages in visualization and visual analytics pipelines, as well as visualization methods~\cite{Lu2017-SOS-9, Lu2017-SOS-10, Wang2016-SOS-13, Zhang2018-SOS-15, Alicioglu2022-SA2020-3}.
Specifically, Wang et al.~\cite{Wang2016-SOS-13} review visual analytics pipelines by examining individual modules from the perspectives of data, visualization, models, and knowledge. Advancing this pipeline concept into the realm of deep learning.
Alicioglu and Sun~\cite{Alicioglu2022-SA2020-3} review visual analytics for XAI, categorizing the current literature based on visual approaches and model usage—such as feature selection, performance analysis, and architecture understanding.
Furthermore, the two works on predictive visual analytics~\cite{Lu2017-SOS-9, Lu2017-SOS-10} discuss relevant techniques on stages such as data pre-processing, feature engineering, model training, and model selection\&validation~\cite{Wang2024-SA2020-1, Yu2018-SOS-14, Hohman2019-SOS-6, Dudley2018-SOS-3}.

Some studies apply miscellaneous or multiple taxonomies, including data perspectives~\cite{Wang2024-SA2020-1} and user types~\cite{Yu2018-SOS-14}.
The notable work by Hohman et al. \cite{Hohman2019-SOS-6} organizes the role of visual analytics in deep learning according to the five Ws (Why, Who, What, When, Where) and How. The ``Where'' dimension refers to the application domain of models, but does not indicate the model space taxonomy used in our work.

Surveys of the fourth group are most relevant to our taxonomy~\cite{Sacha2016-SOS-17, Garcia2018-SOS-5, La2023-SA2020-2, Yuan2021-SA2020-5}.
Sacha et al.~\cite{Sacha2016-SOS-17} propose a pipeline using a machine learning model to analyze data, and then discuss techniques on each stage of the pipeline.
However, this pipeline largely remains data-centric rather than model-centric.
Garcia et al.~\cite{Garcia2018-SOS-5} study the low-level tasks for model visualization, which are related to the model quantification concept of our work.
However, only three types of model quantification, model architecture, training, and features of deep learning models are discussed.
La Rosa et al.~\cite{La2023-SA2020-2} also introduce multiple model-related categorization methods, similar to ours, such as visualization types, analysis methods, and subjects, but their primary category is determined by the explanation method and is essentially not a model-centric taxonomy.
The work by Yuan et al.~\cite{Yuan2021-SA2020-5} is similar to our proposal in that they also proposed the temporal taxonomy that classifies techniques as before/during/after model training.
Beyond that, they do not utilize the model space concept in their classification, which is central to our proposal.

Unlike these related studies, our work introduces a new taxonomy designed from a spatial and temporal perspective.
Moreover, our taxonomy aims to generalize for both traditional machine learning models and modern deep learning models.

\section{MODEL VISUALIZATION FRAMEWORK BY SPATIAL AND TEMPORAL LISTENERS}
\label{sec:framework}

We introduce a model-centric framework that orchestrates ModelVis techniques according to their spatial and/or temporal attachment to the machine learning models interpreted while still maintaining compatibility with the classical InfoVis pipeline.
Here, \textit{space} refers to the abstract spatial extension of a model (e.g., input/intermediate/output data ports, data processing layers, neural modules computing activations, and the overall interconnection topology within a model), while \textit{time} refers to the temporal dynamics of all these spatial elements of a model (e.g., during re-configuration, training, or inference).

As shown in Figure~\ref{fig:framework}, the framework is composed of two stages. In the first stage, the model to be visualized becomes the first-class, central object. The analysis is further conducted through multiple spatial and temporal listeners attached to the model object. The listening outcomes are then translated into relevant, consumable data objects by executing quantification algorithms, e.g., performance index computer. In the second stage, the data objects describing the spatial and temporal behavior of a model are visualized through the classical InfoVis pipeline after customization according to the ModelVis context.


\subsection{Model Listeners}
Model listeners (Figure~\ref{fig:framework}.A) are defined to be the software layer (or AI hardware) embedded inside the machine learning system, which capture the signals related to the model behavior. For example, listeners can report input data, training configuration, model structure, learnable parameters, transient state, output results, and their temporal dynamics (colored-bordered boxes in Figure~\ref{fig:framework}.A). In our framework, the listeners are classified into spatial and temporal categories, focusing on the model's static and dynamic behavior.

From another perspective, the listeners can also be classified as passive model listeners and active model listeners. In most cases, passive listeners are deployed in which the model is agnostic to the listening software/hardware mechanism, so that the model behavior is undisturbed. For some minor cases, e.g. when the full model machinery should be mapped, active model listeners can be applied, so that more model behavior can be exposed and studied. A typical example is the perturbation-based model interpretability method where the input data/feature is manipulated to visualize certain pattern in resulting predictive outcomes (e.g., activation maximization) and/or the input-output correlation.

\subsection{Quantification Algorithms}
To bridge the model evidence from the listening layer and the follow-up visualization and visual analysis, we introduce a quantification algorithm layer which translates the raw signals captured by the model listeners into structured data objects describing the concerning aspect of a machine learning model. The most straightforward example lies in the computation of various performance indicators by comparing the model inference outcomes with their ground-truths. Yet, more complicated quantification algorithms are also essential to understand the sophisticated model behavior. For instance, the deconvolution and layer wise relevance propagation (LRP) quantification algorithms are model-interpretation methods that reverse the forward processing of a model (typically a neural network) to highlight the most important input features with respect to a selected model outcome.


\subsection{Compatibility with the InfoVis Pipeline}
After quantification, listener outputs are organized into standard data objects with multiple types---multi-dimensional (nD) quantitative, one-dimensional (1D) quantitative, relational, temporal, nominal---and other miscellaneous data types. This data abstraction enables the reuse of well-established InfoVis pipeline (Figure~\ref{fig:framework}.B): data, transformation, visualization, and interaction. For example, the transformation operators in InfoVis are then applicable, including filtering, aggregation, projection, distributional summarization, and comparison. Model-centric visualizations can then be decomposed into concurrent, multiple usage of the classical pipeline, which are finally linked together into a visual analytics system, such as correlating output predictive errors with the internal model states, simplifying large model structures, or comparing model performance under different training settings. We note that the data transformation module in the second stage of our framework bears subtle difference from the model quantification algorithms. The quantification algorithm focuses on computing semantically significant representation of a model while the data transformation in InfoVis focuses on preparing structured data for visual representation, i.e., semantic vs. visual.

Specially in the visualization stage, the structured and transformed data objects are mapped to classical visualization genres such as statistical charts, node-link diagrams, parallel coordinates, heatmaps, Sankey diagrams, and other glyphs. The usual ModelVis system combining coordinated multiple views are also supported by our framework as the separate usage of the pipeline and finally linking coordinated analysis together. Users iteratively construct, refine, and diagnose hypotheses through ModelVis systems, and re-route the findings back to the model (listening) layer for updates, forming a closed analysis loop. Therefore, the framework can be seen as not only a ModelVis taxonomy but also a design template fostering new ModelVis techniques and applications.

\section{APPLICATION ON VAST PAPER COLLECTION}
\label{sec:app}

We apply the proposed framework to a large-scale VAST literature corpus.
We first build a high-recall candidate set from publication metadata, then obtain a high-precision ModelVis subset at the paper level, and subsequently perform figure-level framework extraction for cross-paper analysis.

Our method is organized into three stages as detailed in the next section:
\begin{itemize}
  \item \textbf{Stage 1:} paper-level ModelVis screening.
  \item \textbf{Stage 2:} figure-level ModelVis relevance detection with representative-figure selection.
  \item \textbf{Stage 3:} framework-aligned four-dimension extraction, including model listener, data type, visualization type, and visualization purpose.
\end{itemize}

\subsection{Data Collection and Preprocessing}
\label{sec:app:corpus}

Model visualization research is predominantly developed within the model-centric visual analytics community. Therefore, we focus on papers from IEEE VAST (2010---2020), a primary venue for visual analytics studies closely related to model interpretation and diagnosis.
Since 2021, VAST has been integrated into the unified IEEE VIS proceedings, making it difficult to reliably separate VAST papers from the other tracks using metadata alone.
To ensure consistent coverage after this integration, we include all IEEE VIS papers from 2021 to 2024, and then retrieve paper metadata, for example, title, abstract, and keywords when available, from VisPubData~\cite{Vispubdata}, resulting in an initial pool of 1052 papers.

Because a large portion of papers in this pool are unrelated to model visualization, we apply a keyword-based preprocessing step to remove non-ModelVis papers.
Specifically, we manually curate a list of keywords and require each retained paper to contain at least one keyword in its title, abstract, or author-provided keywords. The keyword list includes \textit{model}, \textit{learning}, \textit{analytics}, and \textit{analysis}.
After this filtering step, 514 papers remain for subsequent analyses.

\subsection{LLM-based ModelVis Paper Selection}
\label{sec:app:screening}

An LLM retrieval-augmented paper-classification workflow is designed to classify each paper into ModelVis topic and non-ModelVis topic. Initially, 68 papers are randomly selected and manually determined to be ModelVis-related (35 papers) or unrelated (33 papers). The manually-curated dataset is used as both the evaluation reference and the few-shot example pool for LLM classification.

For each target paper to be classified, we use the BM25 algorithm to retrieve the top-6 neighbors from the labeled pool using title and abstract, while enforcing mixed positive and negative examples in the prompt context. Two LLMs are then queried independently, and a paper is categorized as ModelVis only when both models predict positive.

\begin{figure}
  \centering
  \includegraphics[width=1\linewidth]{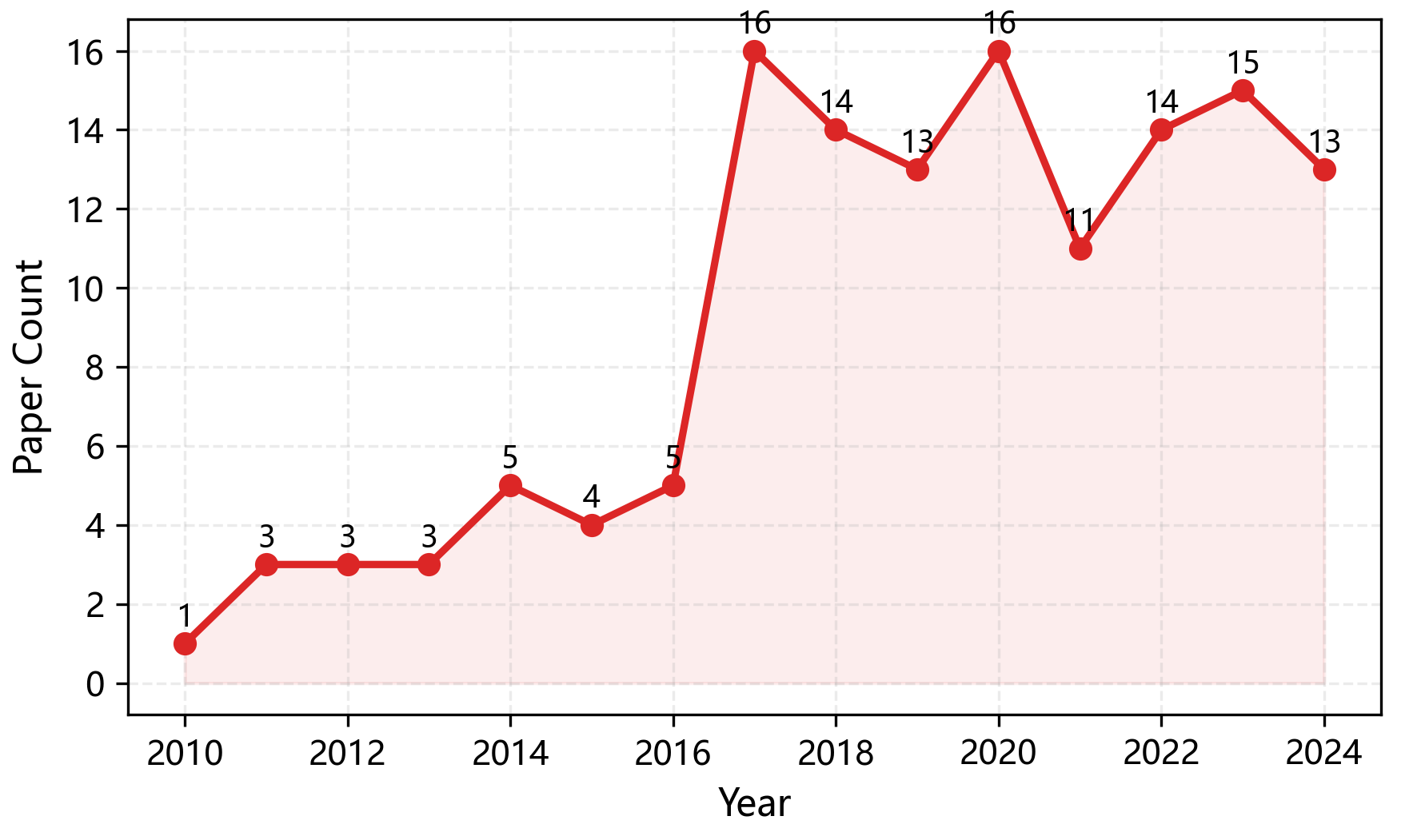}
  \caption{
    Annual distribution of the 136 identified ModelVis papers in our VIS/VAST corpus.
    The red line shows the number of papers per publication year, highlighting a marked increase after 2016 (11-16 papers/year).
  }
  \label{fig:paper-year-all}
\end{figure}

On leave-one-out evaluation over the 68 labeled papers, this method reaches a precision of 0.939 (Table~\ref{tab:stage-performance}).
Applying the same workflow to the 514 candidates yields 136 ModelVis papers (including the 35 manually labeled positives), which form the input for figure-level analysis.
The annual distribution of these 136 papers is shown in Figure~\ref{fig:paper-year-all}.

\subsection{Operational Categories for Human--LLM Extraction}

We select four important dimensions along the ModelVis pipeline to conduct the follow-up study. On each dimension, multiple categories are manually created and then refined during the manual annotation process.
\begin{itemize}
  \item \textbf{Model listener:} input data, training configuration, model structure, learnable parameters, transient state, dynamics (time), output results.
  \item \textbf{Data type:} multi-dimensional quantitative, one-dimensional quantitative, relational, temporal, nominal, and other.
  \item \textbf{Visualization type:} statistical chart, node-link diagram, parallel coordinates, heatmap, Sankey diagram, and others.
  \item \textbf{Visualization purpose:} performance evaluation, I/O relationship, distribution, dimensionality reduction, and other.
\end{itemize}

\subsection{LLM-scaled Model Visualization Learning}
\label{sec:app:stage-2}
We collect full PDF files for the 136 ModelVis papers, converting them into plain-text representations, and then extract figure-level context from the converted text as evidence for downstream retrieval and LLM extractions.
Then, we extend paper-level screening to figure-level framework extractions with a human--LLM workflow.
The objective is to identify key model listeners of each work and map them into a four-field pipeline: model listener, data type, visualization type, and visualization purpose.

\begin{figure}
  \centering
  \includegraphics[width=1\linewidth]{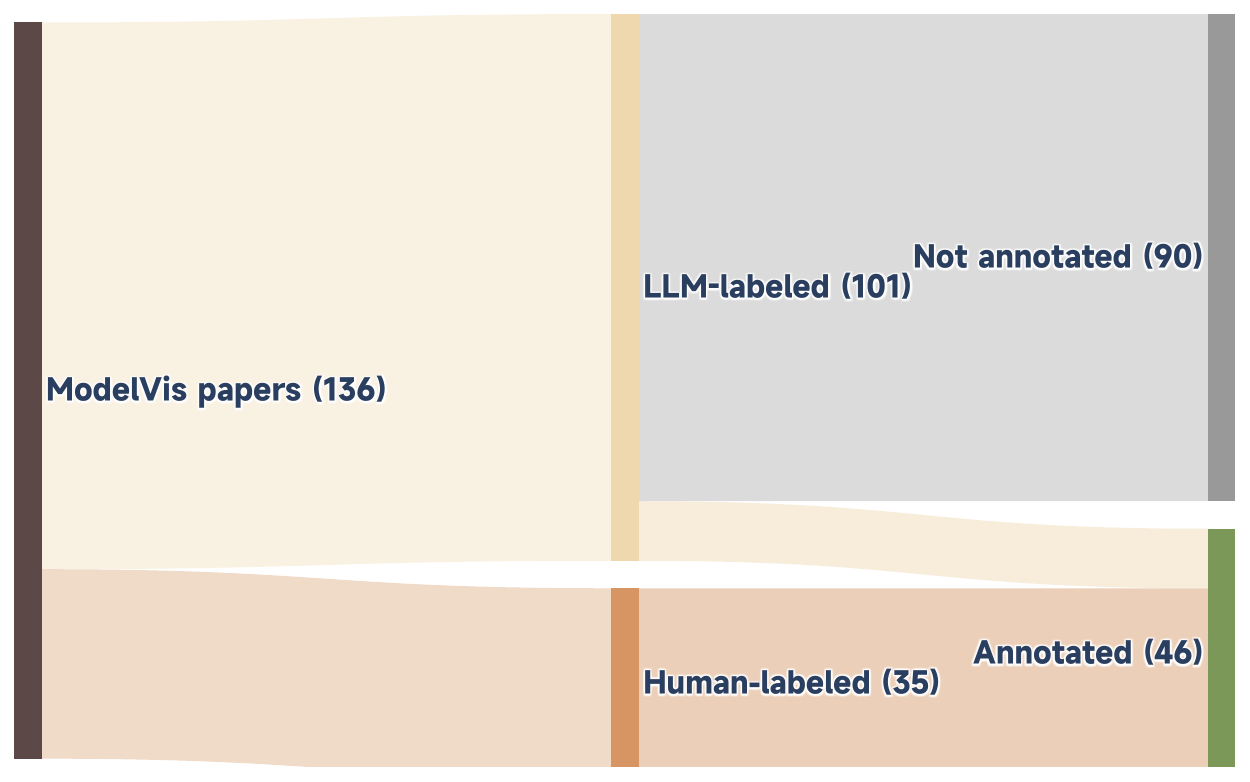}
  \caption{
    Summarizing the sampling of the 136 identified ModelVis papers.
    Papers are first grouped by identification source (human-labeled, $n$ = 35; LLM-labeled, $n$ = 101) and then by whether they were selected for Stage~3 figure-level annotation ($n$ = 46) or not ($n$ = 90).
  }
  \label{fig:dataset-info-sankey}
\end{figure}

The figure-level relevance detection is performed in Stage~2. Using the 46 manually coded papers as samples, we retrieve similar papers by BM25 on title and abstract, construct positive and negative figure exemplars, and classify each target figure as ModelVis-relevant with confidence and short textual evidence.
The sources of these 46 papers are shown in Figure~\ref{fig:dataset-info-sankey}.
To limit the annotation workload, Stage~2 keeps at most three representative figures per paper, prioritizing overview, performance, and mechanism-related evidence.

In Stage~3, each figure selected in Stage~2 is coded for four framework dimensions: model listener, data type, visualization type, and visualization purpose.
For each target figure, we retrieve top-$k$ similar labeled figures from a figure-level BM25 corpus and provide them as in-context examples. Predicted labels are normalized to a controlled vocabulary, and unmatched values are mapped to ``other'' for schema consistency.

\subsection{Results and Analysis}
\label{sec:app:res}

We evaluate the quality of extractions with leave-one-out experiments on the 46 manually coded papers.
Stage-wise quantitative results are summarized in Table~\ref{tab:stage-performance}.
Stage~2 (figure-level relevance detection) achieves an F1 of 0.798 on all folds.
Stage~3 achieves a micro-F1 of 0.848 for model listener, 0.743 for data type, 0.753 for visualization type, and 0.808 for visualization purpose on selected figures with all labels.

\begin{figure*}
  \centering
  \includegraphics[width=1\linewidth]{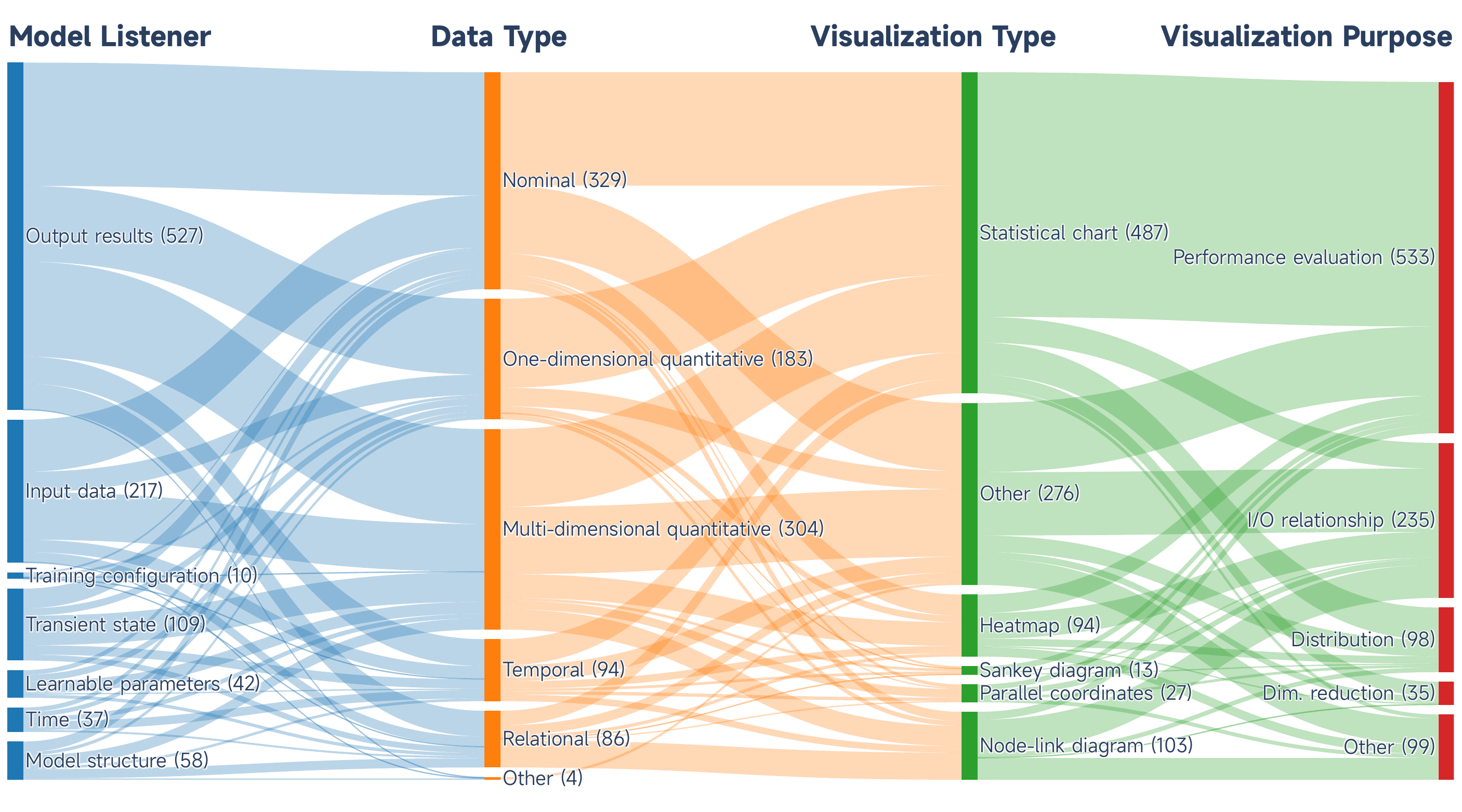}
  \caption{
    A visualization of parallel sets of 1000 ModelVis paths at figure-level for the final 331-figure corpus on  \textit{model listener} $\rightarrow$ \textit{data type} $\rightarrow$ \textit{visualization type} $\rightarrow$ \textit{visualization purpose}.
    Node heights and link widths indicate the relative prevalence of labels and cross-stage transitions.
    The dominant routes are output-oriented listening through quantitative/nominal representations to \textit{statistical chart} and then to \textit{performance evaluation}, with secondary flows toward \textit{I/O relationship} and other visualization types.
  }
  \label{fig:merged-paths}
\end{figure*}

We then run the full pipeline on 90 unlabeled target papers with the same 46-paper sample library.
Stage~2 selects at least one representative figure for 82 papers and outputs 226 selected figures for extractions in Stage~3.
The remaining 8 unlabeled papers have no ModelVis-relevant figure detected in Stage~2 and are therefore excluded from subsequent figure-level analysis.
Combining the 82 retained unlabeled papers with the 46 manually coded papers yields a final analysis set of 128 papers (from the initial 136 ModelVis papers) with at least one ModelVis figure and 331 coded ModelVis figures.
This final 128-paper corpus (2010--2024) also shows a clear growth trend after 2017 in Figure~\ref{fig:paper-year-all}, indicating that ModelVis has evolved from early exploratory efforts to a sustained research stream in recent VIS/VAST publications.

To formalize the transitions within these figures, we derive ModelVis paths by establishing all possible pair-wise connections between labels in adjacent stages; for instance, if a figure is annotated with labels {A, B} in one stage and {C} in the next, two distinct paths (A-C and B-C) are generated. This systematic expansion resulted in a total of 1,000 ModelVis paths extracted from the 331 figures.
By summarizing these ModelVis paths, their distribution is visualized in Figure~\ref{fig:merged-paths}.
This Sankey view reveals a clear dominant chain across all four dimensions: most paths start from \textit{output results} (527), are represented as \textit{nominal} (329) or \textit{multi-/one-dimensional quantitative} data (304/183), then flow to \textit{statistical chart} (487), and finally end at \textit{performance evaluation} (533).
The pattern indicates that current ModelVis studies are primarily organized around result-centric comparison workflows, where structured categorical/quantitative evidence is summarized by chart-based visual encodings for model assessment.
Secondary but non-negligible routes include transitions to \textit{I/O relationship} (235) and \textit{distribution} (98), often through \textit{node-link diagram} (103), \textit{heatmap} (94), and other customized views (276), suggesting that mechanism-oriented and relation-oriented analysis is present but less dominant.

\begin{figure*}
  \centering
  \includegraphics[width=1\linewidth]{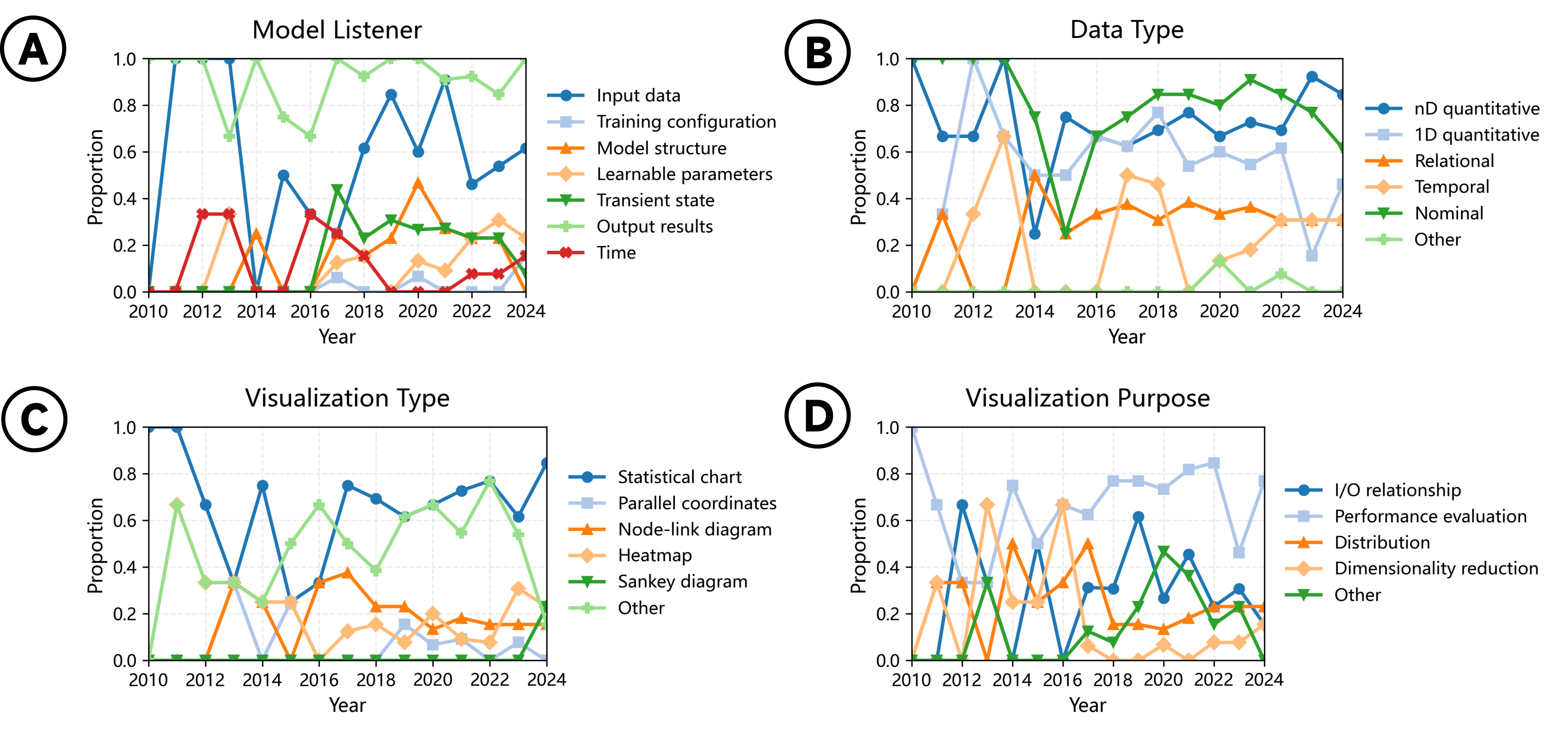}
  \caption{
    Yearly category proportions in the 128-paper ModelVis corpus (2010---2024) across the four framework labels.
    Charts A--D correspond to the model listener, data type, visualization type, and visualization purpose, respectively.
    Each curve reports the within-year proportion of papers assigned to a category, showing sustained dominance of output-results listening and performance-oriented analysis, with a stronger presence of mechanism-related categories after 2017.
  }
  \label{fig:yearly-label-trends}
\end{figure*}

Figure~\ref{fig:yearly-label-trends} further reveals how label usage evolves over time.
In the Model Listener panel (A), \textit{output results} remains the most stable and prevalent target, while mechanism-related listeners (\textit{model structure}, \textit{learnable parameters} and \textit{transient state}) appear more frequently after 2017.
In Data Type (B), \textit{1D/nD quantitative} and \textit{nominal} dominate most years, indicating that both categorical and high-dimensional quantitative evidence are central in ModelVis studies.
In visualization type (C), \textit{statistical chart} remains the major categories throughout the period.
In visualization purpose (D), \textit{performance evaluation} remains the primary objective, while \textit{I/O relationship} and \textit{distribution} analysis maintain visible proportion in later years.
Overall, the temporal pattern indicates a shift from predominantly result-oriented reporting toward broader mechanism- and diagnosis-related analysis.

\begin{figure*}
  \centering
  \includegraphics[width=1\linewidth]{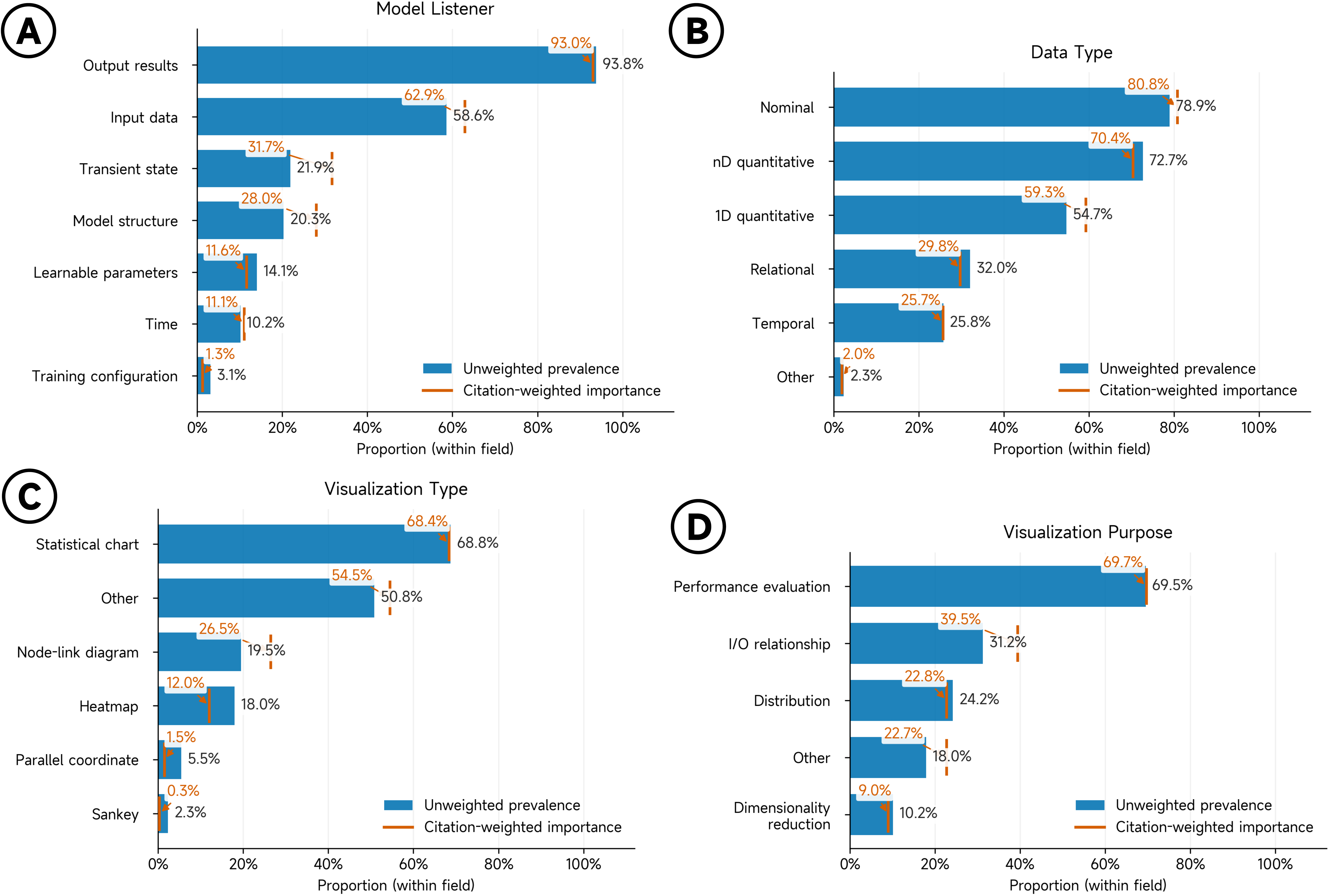}
  \caption{
    Comparison between unweighted prevalence (blur bars) and citation-weighted importance (orange marks) across the four framework labels.
    Charts A--D correspond to model listener, data type, visualization type, and visualization purpose.
  }
  \label{fig:citation-weighted-ranking}
\end{figure*}

As shown in Figure~\ref{fig:citation-weighted-ranking}, at the paper level, the \emph{output results} is the dominant model listener (93.8\%), followed by the \emph{input data} (58.6\%);
\emph{nominal} (78.9\%) and \emph{multi-dimensional quantitative} (72.7\%) are the most frequent structured data types;
\emph{statistical charts} is the most frequent visualization type family (68.8\%);
and \emph{performance evaluation} is the dominant purpose (69.5\%).
These results indicate a prevailing result-centric design pattern in current ModelVis literature.

We additionally collect citation counts for all 128 papers from IEEE Xplore.
Since cumulative citations are conditioned on publication age, a direct cross-year comparison introduces temporal exposure bias toward earlier papers.
To obtain age-adjusted influence estimates, we define the annualized citation weight $w$ for each paper $i$:
\[
  w_i = \frac{\text{citations}_i}{2026 - \text{year}_i + 1}\;.
\]
This normalization yields a per-year citation-intensity proxy and provides a time-standardized weighting scheme, so weighted label proportions reflect relative impact rather than only prevalence.
We then aggregate weighted paper coverage within each label field and obtain citation-weighted category rankings.
The comparison between unweighted prevalence and citation-weighted importance is shown in Figure~\ref{fig:citation-weighted-ranking}.
Notably, \textit{transient state} and \textit{model structure} show stronger citation-weighted importance than their unweighted prevalence, suggesting that mechanism-oriented studies are fewer but comparatively more influential.

\section{RETRIEVAL-AUGMENTED METHOD FOR HUMAN-LLM LITERATURE EXTRACTION}
\label{sec:method}

The details of our literature extraction workflow are explained in this section.
The core challenge is to map a large, weakly structured paper corpus to a framework-defined label space while preserving precision and consistency.
We address this challenge with a retrieval-augmented human--LLM pipeline that progressively refines decisions from the paper level to the figure level, and eventually to framework-aligned multi-label extraction.

\subsection{Problem Definition and Label Space}
\label{sec:method:formulation}

Let $\mathcal{C}$ denote the candidate paper set after dataset preprocessing.
Our goal is to construct a coded subset $\mathcal{M}\subseteq\mathcal{C}$ and assign each selected figure four framework dimensions: model listener, data type, visualization type, and visualization purpose.
The first two are multi-label dimensions, and the latter two are single-label dimensions.

To balance annotation cost and semantic coverage, we decompose the task into three stages as introduced in the last section.
\subsection{Stage 1: Retrieval-augmented Paper-level Screening}
\label{sec:method:stage1}

This stage performs binary classification (ModelVis, $\hat{y}=1$ vs. non-ModelVis, $\hat{y}=0$) using title and abstract. Given a target paper $p$, we retrieve similar labeled papers with the BM25 algorithm over title-and-abstract tokens. We then build a few-shot context from top-ranked neighbors with class-balance constraints (default $k=6$, with minimum positive and negative examples).

Retrieved neighbors are injected into a prompt template and sent to two LLMs (DeepSeek-V3.2 \& ChatGPT-5.1) independently.
We adopt a strict consensus policy:
\begin{equation}
  \hat{y}(p)=1 \iff \hat{y}_{\text{DeepSeek}}(p)=1 \ \wedge\ \hat{y}_{\text{ChatGPT}}(p)=1\;.\nonumber
\end{equation}
Therefore, only papers classified as ModelVis-relevant by both models are retained.
This rule intentionally biases toward high precision, because false positives in Stage~1 could propagate substantial manual burden to subsequent figure-level extraction.

\subsection{Stage 2: Figure-level ModelVis Detection}
\label{sec:method:stage2}

Figure-level extraction requires figure-grounded textual evidence. We therefore convert each selected PDF file into a structured text representation, from which figure captions and surrounding prose can be consistently indexed.

We then extract figure context through keyword-based reference matching in the converted text. Specifically, caption headers and in-text references containing tokens such as ``Fig.'' and ``Figure'' are used to locate figure-specific evidence paragraphs.
The full text is segmented by paragraph, obvious non-body fragments are filtered, and the local context is expanded by retaining the previous and following paragraphs around each direct hit. The final evidence for each figure is formed by combining its caption with this expanded context for subsequent retrieval and prompting.

For each positive paper in Stage~1, we classify whether each figure is ModelVis-relevant in Stage~2.
This stage first retrieves top-$k$ similar labeled papers by title-and-abstract BM25 (default $k=5$), then samples positive and negative figure exemplars from those neighbors. Each target figure is classified with confidence and short evidence snippets.

To control downstream scale and improve representativeness, Stage~2 enforces a top-3 representative figure policy per paper. The retained figures are aligned to overview, performance, and mechanism-oriented roles. Post-processing ensures that each input figure has a valid output entry, confidence scores are clipped to $[0,1]$, and malformed responses are converted to safe defaults.

\subsection{Stage 3: Framework-aligned Four-field Extraction}
\label{sec:method:stage3}

This stage labels each selected figure in Stage~2 on the four dimensions of our framework. We first build a figure-level BM25 corpus from manually coded papers by aligning base figure identities with adjudicated labels. For BM25 indexing and querying, each figure is represented by concatenating its caption repeated three times with its extracted local context.
This caption-upweighting heuristic increases the lexical weight of figure-defining terms and reduces dilution from generic surrounding prose, so that the retrieval is better aligned with figure semantics.
For each target figure, we retrieve top-$k$ similar labeled figures (default $k=10$, with a per-paper cap to avoid source dominance), and feed retrieved examples plus target evidence into the final prompt.

The LLM output is normalized with strict schema checks:
\begin{itemize}
  \item alias mapping for lexical variants,
  \item invalid-value fallback to ``other'',
  \item confidence clipping to $[0,1]$,
  \item evidence-length control for concise auditability.
\end{itemize}

When sub-figures are mapped to the same base figure, we aggregate labels at base-figure granularity: multi-label dimensions are merged by set union, while single-label dimensions are resolved by vote; unresolved conflicts are mapped to ``other''.

\subsection{Evaluation}
\label{sec:method:eval}

We evaluate the figure-level pipeline with leave-one-out over manually extracted papers. In each fold, one paper is held out as a target, and all remaining papers form the retrieval sample library, preventing target leakage in both Stage~2 and Stage~3 retrieval.

Stage~2 is evaluated as a binary classification on explicitly labeled figures only.
Stage~3 is evaluated on figures with available labels. For all multi-label dimensions (model listener and data type), we compute micro-F1 using:
\begin{equation}
  \mathrm{TP}=|Y\cap\hat{Y}|,\quad
  \mathrm{FP}=|\hat{Y}\setminus Y|,\quad
  \mathrm{FN}=|Y\setminus \hat{Y}|.\nonumber
\end{equation}
where $Y$ is the set of true labels and $\hat{Y}$ is the set of predicted labels for a given figure.

The full comparison is summarized in Table~\ref{tab:stage-performance}.
In Stage~1, the BM25 majority-vote baseline yields precision $0.608$, while DeepSeek-V3.2 in 0-shot already reaches $0.933$.
Under 6-shot retrieval augmentation, the single-model runs of DeepSeek-V3.2 and ChatGPT-5.1 both obtain precision $0.889$, and the dual-model consensus reaches the best precision of $0.939$.

In Stage~2, the 0-shot setting obtains F1 $0.722$, and 5-shot improves it to $0.798$ (+$0.076$).
In Stage~3, 10-shot consistently outperforms 0-shot across all four dimensions: model listener ($0.848$ vs. $0.788$), data type ($0.743$ vs. $0.706$), visualization type ($0.753$ vs. $0.525$), and visualization purpose ($0.808$ vs. $0.754$).

\begin{table*}[ht]
\centering
\caption{
Stage-wise comparative performance under leave-one-out evaluation.
The table includes Stage~1 paper-level screening baselines (BM25 majority vote, 0-shot, and 6-shot LLM settings), Stage~2 figure-level detection (0-shot vs.\ 5-shot), and Stage~3 four-field extraction (0-shot vs.\ 10-shot).
TP/FP/TN/FN are aggregated across folds; reported scores are precision (Stage~1), F1 (Stage~2), and micro-F1 (Stage~3).
}
\label{tab:stage-performance}

\begin{tabular}{@{}ccccccccc@{}}
\toprule
\multirow{2}{*}{Stage} &
  \multirow{2}{*}{Method} &
  \multirow{2}{*}{Model / Algorithm} &
  \multirow{2}{*}{Target} &
  \multicolumn{4}{c}{Metric} &
  \multirow{2}{*}{Reported score} \\ \cmidrule(lr){5-8}
 &
   &
   &
   &
  TP &
  FP &
  TN &
  FN &
   \\ \midrule
\multirow{5}{*}{Stage 1} &
  majority vote &
  BM25 &
  \multirow{5}{*}{\begin{tabular}[c]{@{}c@{}}manually classified\\ 68 papers\end{tabular}} &
  31 &
  20 &
  13 &
  4 &
  Precision = 0.608 \\ \cmidrule(lr){3-3} \cmidrule(l){5-9} 
 &
  0-shot &
  DeepSeek-V3.2 &
   &
  28 &
  2 &
  31 &
  7 &
  Precision = 0.933 \\ \cmidrule(lr){3-3} \cmidrule(l){5-9} 
 &
  \multirow{3}{*}{6-shot} &
  DeepSeek-V3.2 &
   &
  32 &
  4 &
  29 &
  3 &
  Precision = 0.889 \\
 &
   &
  ChatGPT-5.1 &
   &
  31 &
  2 &
  31 &
  4 &
  Precision = 0.889 \\
 &
   &
  \begin{tabular}[c]{@{}c@{}}ChatGPT-5.1\\ \& DeepSeek-V3.2\\ consensus\end{tabular} &
   &
  31 &
  2 &
  33 &
  2 &
  \textbf{Precision = 0.939} \\ \cmidrule(l){2-9} 
\multirow{2}{*}{Stage 2} &
  0-shot &
  \multirow{2}{*}{DeepSeek-V3.2} &
  \multirow{2}{*}{\begin{tabular}[c]{@{}c@{}}46 labeled\\ ModelVis papers\end{tabular}} &
  61 &
  3 &
  / &
  44 &
  F1 = 0.722 \\
 &
  5-shot &
   &
   &
  73 &
  5 &
  / &
  32 &
  \textbf{F1 = 0.798} \\ \cmidrule(l){2-9} 
\multirow{8}{*}{Stage 3} &
  \multirow{4}{*}{0-shot} &
  \multirow{8}{*}{DeepSeek-V3.2} &
  Model Listener &
  82 &
  19 &
  / &
  25 &
  micro-F1 = 0.788 \\
 &
   &
   &
  Data Type &
  77 &
  21 &
  / &
  43 &
  micro-F1 = 0.706 \\
 &
   &
   &
  Visualization Type &
  32 &
  29 &
  / &
  29 &
  micro-F1 = 0.525 \\
 &
   &
   &
  Visualization Purpose &
  46 &
  15 &
  / &
  15 &
  micro-F1 = 0.754 \\ \cmidrule(l){4-9} 
 &
  \multirow{4}{*}{10-shot} &
   &
  Model Listener &
  106 &
  16 &
  / &
  22 &
  \textbf{micro-F1 = 0.848} \\
 &
   &
   &
  Data Type &
  101 &
  27 &
  / &
  43 &
  \textbf{micro-F1 = 0.743} \\
 &
   &
   &
  Visualization Type &
  55 &
  18 &
  / &
  18 &
  \textbf{micro-F1 = 0.753} \\
 &
   &
   &
  Visualization Purpose &
  59 &
  14 &
  / &
  14 &
  \textbf{micro-F1 = 0.808} \\ \bottomrule
\end{tabular}
\end{table*}

\section{DISCUSSION}
\label{sec:discussion}

The LLM-based experiment of this work demonstrates the validity of the proposed ModelVis framework. Most papers in our studied collection relating to model visualization can be mapped to the framework with either spatial or temporal listeners extracted. The follow-up visualization mostly follows the classical InfoVis pipeline.

\subsection{Dissimilarity with Data Visualization}

We also identify several differences on the state-of-the-art model visualization research, in comparison to the classical visualization study focusing on data. First, among ModelVis papers in our collection, there is few novel visual metaphor design, which had been the core of traditional visualization research. The reason probably lies in that the machine learning models are already complex to comprehend, introducing new visualization design may further increases the learning difficulty and reduce usability. Second, the presents of temporal listeners and temporal data visualizations are still minority in ModelVis studies (Figure 4). We ascribe this to the relatively larger effort required to measure model dynamics in comparison to their static features. Yet, visualizing model dynamics can be a promising, under-exploited topic. Finally, unlike the visual analytics research highly concerning the analytics process, existing ModelVis proposals focus more on key dimensions of a model, mostly its output results and input data, as well as the input-output correlations. Delineating the overall model functioning process calls for interdisciplinary study, by joint-forcing visualization and AI research communities.

\subsection{Trends on Model Visualization Research}

Our study also unveils multiple trends on ModelVis research. The year-by-year curve on number of papers in Figure 2 shows the surge of the topic from 2017 in visualization community, the hotness of which still goes on now. Despite its popularity, we observe that visualization works on deeper mechanism of machine learning models, e.g., model structure, transient states, and input/output relationship, drop in recent years (Figure 5). A re-invent of ModelVis research can be essential currently to continue to open the black-box of highly valuable AI systems. Finally, on the data types listened from machine learning models, the multi-dimensional data still dominates (Figure 5.B), which renders the importance enduring significance of multi-dimensional data visualization in ModelVis research.

\subsection{Limitation and Future Work}

The work presented here does have certain limitations due to its broad coverage. First, the current categories of ModelVis framework dimensions can be non-exhaustive, and parts of the extraction process may be affected by subjective judgment. For example, we list seven spatial and temporal model listeners through the discussion among co-authors and paper annotators. Conducting more user studies within the visualization and AI community, especially with machine learning practitioner, can help to enrich the classification of our ModelVis framework. Second, the framework is now applied to the ModelVis papers within the IEEE VIS collection, hence our result is influenced by the selected dataset and its coverage. Expanding the study to all visualization venues as well as AI events related to model visualization will further generalize the result of this work and validate our framework more extensively. Finally, from the application perspective, the theoretic ModelVis framework ought to be deployed as software libraries enclosing all state-of-the-art model visualization techniques, facilitating the actual usage and implementation of ModelVis research.

\section{CONCLUSION}
\label{sec:conclusion}

We revisit model visualization through a model-centric, spatial, and temporal perspective, organizing prior work into an InfoVis pipeline-inspired framework.
We describe how abstract listeners can capture essential model behavior, how this model behavior information is translated and structured into analyzable data types, transformed into visualization-friendly representations, and finally mapped to visualization designs and analysis purposes.
We further demonstrate how the framework can be operationalized via a reproducible extraction workflow on a core ModelVis paper collection, combining manual annotation with LLM-assisted augmentation.

Our results and analysis suggest that the current ModelVis research concentrates on recurring patterns, including a strong focus on output-oriented listening, quantitative/nominal data organization, and purpose settings centered on performance evaluation and I/O relationship, while studies on deeper internal model mechanism remain comparatively sparse.
In particular, richer treatments of spatiotemporal coupling, systematic analysis of training configuration and data bias sources, and more complete interaction loops that connect diagnosis back to concrete model/data edits remain promising opportunities.
Beyond forming taxonomy, our framework can serve as a design checklist for model builders, a positioning tool for researchers, and a scaffold for future literature organization.

\section{ACKNOWLEDGMENTS}

This work was supported by NSFC Grant 62572026,  National Social Science Fund of China 22\&ZD153, State Key Laboratory of Complex \& Critical Software Environment (SKLCCSE).

Lei Shi, Yingchaojie Feng, Liang Zhou are the corresponding authors of this paper.

\bibliographystyle{IEEEtran}
\bibliography{refs}

\begin{IEEEbiography}{Siyu Wu}
  is a Ph.D. student at School of Computer Science \& Engineering, Beihang University at Beijing, 100191, China. His research interests include visualization, human-computer interaction, and artificial intelligence. Wu received his Bachelor's degree in Information Engineering from Nanjing University of Information Science \& Technology. Contact him at siyuw@buaa.edu.cn.
\end{IEEEbiography}

\begin{IEEEbiography}{Lei Shi}
  is a Professor in the School of Computer Science and Engineering, Beihang University, 100191, China.
  His current research interests are Data Mining, Visual Analytics, and AI, with more than 100 papers published in top-tier venues.
  He holds B.S. (2003), M.S. (2006) and Ph.D. (2008) degrees from Department of Computer Science and Technology, Tsinghua University.
  He is the recipient of IBM Research Division Award on ``Visual Analytics'' and the IEEE VAST Challenge Award twice in 2010 and 2012. He has organized several workshops on combining visual analytics and data mining and served on the (senior) program committees of many related conferences.
  He is an IEEE senior member.
  He is one of the corresponding authors of this article.
  Contact him at leishi@buaa.edu.cn.
\end{IEEEbiography}

\begin{IEEEbiography}{Lei Xia}
  is a Ph.D. student at School of Computer Science \& Engineering, Beihang University at Beijing, 100191, China. His research interests include visualization, human-computer interaction, and artificial intelligence. Xia received his Bachelor's degree in Computer Science from Beihang University. Contact him at lei\_xiaaa@buaa.edu.cn.
\end{IEEEbiography}

\begin{IEEEbiography}{Cenyang Wu}
  is a Ph.D. student at the National Institute of Health Data Science, Peking University at Beijing, 100191, China. His research interests include scientific visualization, computer graphics, and generative models. Wu received his bachelor's degree in Mathematics and Applied Mathematics from Beijing Normal University.  Contact him at wuceny@stu.pku.edu.cn.
\end{IEEEbiography}

\begin{IEEEbiography}{Zipeng Liu}
  is an Associate Professor at School of Software, Beihang University, Beijing, 100191, China. His research interests include visualization, human-computer interaction, and interpretable AI. Liu received his Ph.D. in Computer Science from University of British Columbia. Contact him at zipeng@buaa.edu.cn.
\end{IEEEbiography}

\begin{IEEEbiography}{Yingchaojie Feng}
  is a research fellow at National University of Singapore at Singapore. His research interests include natural language processing, data visualization, and human computer interaction. Feng received his Ph.D. degree in visualization from Zhejiang University.
  He is one of the corresponding authors of this article.
  Contact him at feng.y@nus.edu.sg.
\end{IEEEbiography}

\begin{IEEEbiography}{Liang Zhou}
  is an Assistant Professor at the National Institute of Health Data Science, Peking University, Beijing, 100191, China. His research interests include visualization, visual analysis, and extended reality for medicine. Dr. Zhou received his Ph.D. in Computing from the University of Utah.
  He is one of the corresponding authors of this article.
  Contact him at zhoulng@pku.edu.cn.
\end{IEEEbiography}

\begin{IEEEbiography}{Wei Chen}
  is a Professor with the State Key Laboratory of CAD\&CG, Zhejiang University, Hangzhou, 310058, China.
  His research interests include visualization and visual analytics.
  He has performed research in visualization and visual analysis and published more than 100 IEEE/ACM Transactions and CCF-A papers.
  He actively served in many leading conferences and journals.
  More information can be found at: {http://www.cad.zju.edu.cn/home/chenwei/}.
\end{IEEEbiography}

\end{document}